\documentclass[pdflatex,sn-mathphys-num]{sn-jnl}

\usepackage{graphicx}%
\usepackage{multirow}%
\usepackage{amsmath,amssymb,amsfonts}%
\usepackage{amsthm}%
\usepackage{mathrsfs}%
\usepackage[title]{appendix}%
\usepackage{xcolor}%
\usepackage{textcomp}%
\usepackage{manyfoot}%
\usepackage{booktabs}%
\usepackage{algorithm}%
\usepackage{algorithmicx}%
\usepackage{algpseudocode}%
\usepackage{listings}%
\usepackage{booktabs}
\usepackage{pifont}
\usepackage{threeparttable}
\usepackage{makecell}
\usepackage{subcaption}
\usepackage[dvipsnames]{xcolor}
\usepackage{tabularx}
\usepackage[labelsep=quad]{caption} 

\theoremstyle{thmstyleone}%
\theoremstyle{thmstyletwo}%

\theoremstyle{thmstylethree}%

\newcommand{\methodname}{HASTE}
\newcommand{\cmark}{\textcolor{ForestGreen}{\ding{51}}}
\newcommand{\xmark}{\textcolor{red}{\ding{55}}}

\newcommand{\norm}[1]{\left \lVert #1 \right \rVert}
\def\sR{{\mathbb{R}}}
\def\sN{{\mathbb{N}}}

\algrenewcommand\algorithmicrequire{\textbf{Input:}}
\algrenewcommand\algorithmicensure{\textbf{Output:}}

\begin{document}

\title[]{\parbox{\textwidth}{\centering HASTE: A Framework for Training-Free, \\
	Dynamic, and Steerable Compression of \\ Pre-Trained Convolutional Neural Networks}}


\author*[1,2]{\fnm{Lukas} \sur{Meiner}}\email{Lukas.Meiner@bosch.com}

\author[1]{\fnm{Jens} \sur{Mehnert}}

\author[1,2]{\fnm{Alexandru Paul} \sur{Condurache}}

\affil[1]{\orgname{Robert Bosch GmbH}, \orgaddress{\city{Leonberg}, \country{Germany}}}

\affil[2]{\orgname{Universit\"at zu L\"ubeck}, \orgaddress{\city{L\"ubeck}, \country{Germany}}}

\abstract{
	Deploying large convolutional neural networks (CNNs) on resource-constrained devices is challenging due to their high computational cost. While dynamic execution methods are promising, existing approaches for CNNs typically require specialized training or fine-tuning, limiting their effectiveness when applied to pre-trained models and requiring data access. To address this gap, we propose \methodname\ (\textbf{Has}hing for \textbf{T}ractable \textbf{E}fficiency), a plug-and-play convolution module that enables training-free, dynamic compression of large pre-trained CNNs. At inference time, \methodname\ uses locality-sensitive hashing to identify and merge redundant channels of latent feature maps on a patch-wise basis. This process simultaneously compresses the depth of both input features and their corresponding filters, resulting in computationally cheaper convolutions. We conduct extensive experiments on CIFAR-10 and ImageNet across a range of architectures, demonstrating a 46.2\% FLOPs reduction in a ResNet34 on CIFAR-10 with only a 1.25\% drop in accuracy, without any retraining. We support our claims by comprehensive ablation studies to validate our core design choices, an analysis of the method's properties and limitations, and a discussion that connects our channel merging scheme to the conceptually related task of token merging in Vision Transformers. Our results demonstrate that \methodname\ provides an effective solution for steerable compression of pre-trained CNNs at runtime, opening new possibilities for the deployment of efficient deep learning methods.
}

\keywords{Steerable model compression, Dynamic execution, Locality-sensitive hashing, Training-free, Convolutional neural networks}



\maketitle

\section{Introduction}
\label{sec:introduction}
	\begin{figure*}[h!]
		\begin{center}
			\includegraphics[width=1\linewidth]{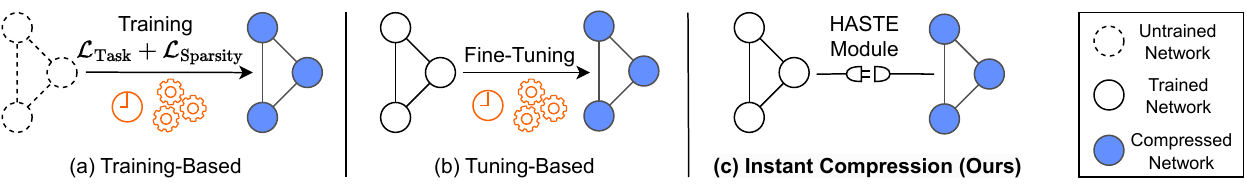}
		\end{center}
		\caption{
			Overview of related compression approaches. Existing methods require training with specialized sparsity losses (a) or fine-tuning to restore accuracy after the compression step (b). Our method (c) instantly compresses the network and maintains performance entirely without training. Adapted from \cite{Meiner2025Data}
		}
		\label{fig:overview}
	\end{figure*}

	With the widespread availability of powerful deep learning hardware, it has become feasible to train increasingly larger models, enabling significant breakthroughs in computer vision tasks. Concurrently, concerns associated with computational requirements of these models, notably the high energy consumption and environmental impact, have grown considerably. These factors are particularly critical at the scale of automotive applications, where models are potentially deployed across millions of vehicles.
	
	In response to these challenges, a variety of carefully crafted efficient architectures have been proposed \cite{Sandler2018MobileNetV2, Tan2021EfficientNetV2, Ma2018ShuffleNet}, tailored explicitly for embedded applications. These models achieve computational efficiency by integrating inductive biases into their architecture. Despite these developments, more scalable architectures \cite{Simonyan2015Very, He2016Deep, Dosovitskiy2021Image} remain popular due to their superior performance and ability to generalize across diverse tasks and domains, despite the higher resource demands in terms of training time, data, and energy.
	
	To balance the trade-off between performance and computational cost, considerable research has emerged in model compression, including 
	unstructured \cite{Wimmer2021COPS, Han2016Deep, Wimmer2022Interspace}
	and structured pruning \cite{Anwar2017Structured, Li2017Pruning, Yeom2021Pruning, Liu2017Learning, He2019Filter, Zhuang2018Discrimination, Xu2021Efficient}, 
	dynamic routing \cite{Cakaj2024CNN, Bejnordi2020Batch, Hua2019Channel, Verelst2020Dynamic, Li2021Dynamic, Liu2019Dynamic, Elkerdawy2022Fire}, 
	quantization \cite{Meiner2025PROM, Kim2022BASQ, Zhu2020Towards}, 
	mixtures of experts \cite{Berisha2025Efficient, Shazeer2017Outrageously, Fedus2022Switch, Belcak2023Fast},
	and knowledge distillation \cite{Hinton2015Distilling, Gou2021Knowledge}.
	Structured pruning, in particular, has garnered significant attention due to its direct resource savings. However, two practical limitations restrict the broader application of existing pruning methodologies, as shown in Figure~\ref{fig:overview}. Firstly, many pruning approaches \cite{Dong2017More, Liu2017Learning, Gao2019Dynamic, Verelst2020Dynamic, Bejnordi2020Batch, Li2021Dynamic, Xu2021Efficient} depend on actively learning which channels to prune during training. This increases the complexity of the optimization process through additional parameters and supplementary loss functions, and requires training the model from scratch to achieve any reduction in inference cost. Secondly, approaches operating on pre-trained models typically require fine-tuning after pruning to restore performance \cite{Wen2016Learning, Li2017Pruning, Zhuang2018Discrimination, He2018AMC}, requiring access to the original training data to prevent catastrophic forgetting \cite{Goodfellow2014Empirical}.
	
	In addressing these limitations, we propose \methodname\ (Hashing for Tractable Efficiency), a plug-and-play convolutional module designed for training-free and dynamic compression of CNN architectures. \methodname\ leverages locality-sensitive hashing (LSH) \cite{Indyk1998Approximate} to identify and cluster redundant channels dynamically within latent feature representations. By exploiting the distributive property of convolutions, \methodname\ efficiently compresses input and filter channels by merging approximately similar ones, substantially reducing the number of floating-point operations (FLOPs) required at inference. The resulting compression ratio and performance trade-off can be directly controlled via a single hyperparameter, simplifying experimentation and deployment.
	
	Our experiments demonstrate that \methodname\ significantly reduces the computational demands of various pre-trained CNNs, maintaining high accuracy without additional training or fine-tuning. Crucially, our method does not require access to the original training dataset, making it suitable for scenarios with strict data privacy and availability constraints. Furthermore, the adaptability of our approach addresses common restrictions in edge computing, such as limited energy budgets, thermal management requirements, and varying computational resources. To our knowledge, \methodname\ represents the first entirely training-free, dynamically adaptable CNN compression method, enabling real-time adjustments to model complexity based on hardware availability.

	This manuscript is an extended and revised version of our VISAPP 2025 conference paper \cite{Meiner2025Data}, expanding on theoretical foundations, additional experiments and ablations, as well as adding broader contextual analysis and conceptual links to token compression methods in Vision Transformers \cite{Dosovitskiy2021Image}.

	Our main contributions can be summarized as follows:
	\begin{itemize}
		\item We identify locality-sensitive hashing as a suitable approach for dynamically identifying structural redundancies in latent features of CNNs, incurring minimal computational overhead and not relying on any training or calibration data.
		\item Based on these findings, we propose \methodname, a plug-and-play replacement for convolutional modules, enabling steerable FLOPs reduction at test time without any training requirements.
		\item We evaluate our method across popular CNN architectures and benchmark vision datasets. We expand on our conference paper by providing extensive ablations for our design choices, additional experiments and analysis, and a discussion on the limitations of our method.
		\item Additionally, we contextualize our contributions with emerging trends, such as efficiency-oriented CNN architectures and token reduction techniques for Vision Transformers, highlighting both challenges and opportunities for future work.
	\end{itemize}

\section{Background And Related Work}
	Model compression has long been an active area of research, especially as the size and computational demands of deep neural networks continue to grow. While a wide range of compression techniques have emerged, our focus lies on the dynamic compression of latent features at inference time. This field closely aligns with structured pruning and dynamic gating approaches, which aim to eliminate structural components such as filters or feature channels, resulting in hardware-friendly speed-ups. 
	
	Recently, the idea of dynamically compressing intermediate representations has also gained traction in the context of Vision Transformers (ViTs) \cite{Dosovitskiy2021Image}, where features take the form of token embeddings. In this domain, techniques like token pruning and token merging reduce the number of tokens passed through subsequent layers, reducing the computational load based on input redundancy.
	
	Among these diverse approaches to model compression, a key consideration lies in whether training or fine-tuning is required to utilize the method effectively. The ability to compress models without any additional training steps is attractive to practitioners and researchers alike, as it enables the use of large, publicly available pre-trained models even on constrained hardware platforms.
	
	\bmhead{Static Pruning and Dynamic Gating}
	Traditional pruning approaches often rely on static criteria to determine which components of a network can be removed. These methods typically require fine-tuning after pruning to restore model performance, or train a model from scratch with additional losses. For example, PFEC \cite{Li2017Pruning} prunes filters based on their $L^1$ norm in a one-shot fashion, while DCP \cite{Zhuang2018Discrimination} equips models with multiple loss terms before fine-tuning to promote highly discriminative channels to be formed. Approaches like Network Slimming \cite{Liu2017Learning} and DMCP \cite{Xu2021Efficient} introduce additional sparsity-promoting losses during model training to facilitate the selection of structures which are suitable for pruning. 
	
	In contrast to static compression, dynamic gating approaches \cite{Li2021Dynamic, Elkerdawy2022Fire, Cakaj2024CNN, Hua2019Channel, Bejnordi2020Batch} allow for input-dependent paths through the network, selectively using structural components dependent on active gates. However, these gating modules must be trained alongside the model, limiting their applicability to pre-trained architectures.  
	
	A handful of approaches eliminate the need for access to the training dataset entirely, but either use synthetic data to retrain the model \cite{Yin2020Dreaming} or generate a static model \cite{Yvinec2023RED, Bai2023Unified} that is unable to adapt its compression to the availability of hardware resources or the level of redundancy in the input dynamically. Our proposed method targets on-the-fly compression of models at inference time.
	
	\bmhead{Hashing-Based Compression} Locality-sensitive hashing \cite{Indyk1998Approximate, Achlioptas2003Database} has found increasing use in efficient inference pipelines, particularly for high-dimensional data. Reformer \cite{Kitaev2020Reformer} applies LSH to attention mechanisms to reduce complexity in Transformers \cite{Vaswani2017Attention}. SLIDE and MONGOOSE \cite{Chen2020SLIDE, Chen2021MONGOOSE} apply LSH to feedforward networks, selectively activating only a subset of neurons for each input. \citet{Mueller2022} extend this concept to neural radiance fields using multiresolution hash encodings.
	
	Other (approximate) nearest neighbor search approaches have also been explored for model compression, such as count sketches to approximate forward passes in multilayer perceptrons (MLPs) \cite{Liu2021Efficient}, or $k$-means clustering for redundancy detection in CNN input channels \cite{Liu2021More}. However, these efforts result in static models or are limited to fixed pruning ratios. In contrast, our method integrates LSH directly into the convolution operation to detect and compress redundant channels dynamically at runtime, without any training or fixed compression target.

	\bmhead{Token Reduction in Vision Transformers} 
	A particularly promising approach for lowering inference cost in Vision Transformers \cite{Dosovitskiy2021Image} is reducing the number of tokens processed throughout the network. This form of dynamic feature compression reduces both the computational cost of self-attention layers and the subsequent MLP blocks, helping to alleviate the quadratic complexity of attention mechanisms. Two dominant strategies have emerged: token pruning \cite{Yin2022ViT, Rao2021DynamicViT, Liang2022Not}, which removes uninformative tokens entirely, and token merging \cite{Chen2023DiffRate, Lee2025Lossless, Kim2024Token, Bolya2023Token}, which combines similar token embeddings into a single representation. Notably, both approaches have demonstrated strong empirical performance and can, in some cases, function without requiring additional training \cite{Liang2022Not, Bolya2023Token, Lee2025Lossless}. 
	
	More recent advances have introduced input-adaptive techniques that control the number of tokens merged based on token similarity thresholds \cite{Lee2025Lossless}, enabling finer control over the computational budget in response to input complexity. We will return to these strategies in Section~\ref{sec:discussion} to highlight both their conceptual overlap with and differences from our proposed method.
	
\section{Methodology}
\label{sec:methodology}
	In this section, we present \methodname\, a plug-and-play convolutional module that reduces inference-time computational cost through a training-free compression mechanism, leveraging locality-sensitive hashing for approximate similarity search. The section begins with a theoretical introduction to LSH. We then describe how it is used to identify redundancies in latent CNN features, and follow with an overview of the HASTE module and its integration into convolutional model architectures. We conclude with the presentation of a design choice that further increases the efficiency of our method. 
	
	\subsection{Locality-Sensitive Hashing via Random Projections}
	\label{sec:method:sub:lsh}
		Locality-sensitive hashing is a probabilistic way for performing efficient approximate nearest neighbor search in high-dimensional spaces. The key property of LSH is that two similar inputs are mapped to the same hash bucket with high probability, while dissimilar inputs are unlikely to collide. This is in contrast to regular hashing schemes, which try to reduce hash collisions to a minimum by widely scattering the input data across hash buckets. More formally, a family of hash functions $\mathcal{H}=\{h: \sR^d \to \sN\}$ is called $(r_1, r_2, p_1, p_2)$-sensitive, if for any two vectors $x, y \in \sR^d$, we have that:
		\begin{equation}
			\begin{aligned}
			&\bullet \text{if } Sim(x,y) \geq r_1, \text{then } \mathbb{P}[h(x)=h(y)] \geq p_1, \\
			&\bullet \text{if } Sim(x,y) \leq r_2, \text{then } \mathbb{P}[h(x)=h(y)] \leq p_2,
			\end{aligned}
		\end{equation}
		where we require $p_1 > p_2$ and $r_1 > r_2$ to hold \cite{Indyk1998Approximate, Chen2020SLIDE}. 
		
		For use in neural network architectures, cosine similarity is a particularly suitable measure. The dot product between an arbitrary row $W_i \in \sR^d$ of a weight matrix $W$ and an input vector $x \in \sR^d$, which is a fundamental operation in both convolutional and fully connected layers, can be expressed as:
		\begin{equation}
			\label{eq:dot_product_cosine_sim_relation}
			W_i \cdot x = \left \lVert W_i \right \rVert \left \lVert x \right \rVert \cos\theta,
		\end{equation}
		where $\theta$ is the angle between $W_i$ and $x$, and $\cos\theta$ represents their cosine similarity. 
		
		A common LSH family for cosine similarity is based on random hyperplanes, also called random projections (RP) \cite{Gionis1999Similarity, Chen2020SLIDE}. By generating a set of $L$ random hyperplanes in $d$-dimensional space, we partition the input space into at most $2^L$ hash buckets. The position of an input vector $x \in \sR^d$ relative to the $l$-th hyperplane, defined by the hyperplane's normal vector $v_l \in \sR^d,\, l \in \{1,\dots,L\}$, is determined by:
		\begin{equation}
			h_l: \sR^d \to \{0,1\},\,\,\, h_l(x) := \begin{cases}
				1,\, \text{if } v_l \cdot x > 0,\\
				0,\, \text{else.}
			\end{cases}
		\end{equation}
		We can construct $v_l$ by sampling its components independently from a standard normal distribution $\mathcal{N}(0,1)$.
		Each $h_l$ returns a binary decision, representing whether the input $x$ lies above ($h_l(x)=1$) or below ($h_l(x)=0$) the $l$-th hyperplane. 
		By concatenating the $L$ binary decisions of all hyperplanes, we receive the hash function:
		\begin{equation}
			h:\sR^d \to \{0,1\}^L,\,\,\, h(x) = \left(h_1(x), \dots, h_L(x)\right).
		\end{equation}
		This $L$-bit binary code acts as a unique identifier of a single hash bucket, and can equivalently be transformed into an integer:
		\begin{equation}
			\begin{gathered}
				h:\sR^d \to \left\{0, \dots, 2^L-1\right\}, \\
				h(x) = 2^{L-1}h_L(x) + \dots + 2^0h_1(x).
			\end{gathered}
		\end{equation}
		The computation of $L$ dot products with the normal vectors of random hyperplanes allow us to effectively partition high-dimensional spaces into distinct identifiable regions, in which points are similar with respect to the chosen similarity measure.

	\subsection{Finding Redundancies with LSH}		
		\label{sec:method:subsec:finding_redundancies}
		Having established LSH with random projections as an efficient method for grouping vectors by cosine similarity, we now detail its application to identify redundant computations within convolutional neural networks. 
		
		A standard convolutional layer processes an input feature map $X \in \sR^{C_\text{in} \times H \times W}$ with a set of learned filters $F_j \in \sR^{C_\text{in} \times K \times K},\, j \in \{1, \dots, C_\text{out}\}$. We denote the input and output channel dimensions as $C_{in}$ and $C_{out}$, respectively, the kernel size as $K$, and the spatial dimensions of the input as $H$ and $W$. At a specific spatial location, the output $Y_j$ of the $j$-th filter is computed by convolving the filter with the corresponding input patch. For a given input channel $c \in \{1, \dots, C_\text{in}\}$, this involves the dot product between the filter's $c$-th channel slice $F_{j,c}$ and the input's $c$-th channel slice $X_c$, both flattened to vectors. Then, we take the sum of contributions from all $C_\text{in}$ channels. Therefore, the output $Y_j$ at that location is:
		\begin{equation}
			\label{eq:convolution_expressed_as_dot_product}
			Y_j = \sum_{c=1}^{C_\text{in}} F_{j,c} \cdot X_{c}.
		\end{equation}
		Using Equation~\ref{eq:dot_product_cosine_sim_relation}, this can be rewritten as:
		\begin{equation}
			Y_j = \sum_{c=1}^{C_\text{in}} \norm{F_{j,c}} \norm{X_{c}} \cos\theta_c \,,
		\end{equation}
		where $\theta_c = \angle(F_{j,c}, X_c)$. This formulation highlights that the output depends on the cosine similarity between filter channel slices and corresponding input channel slices.
		
		Our core idea is that if multiple input channel slices $X_c$ are highly similar in terms of their orientation (i.e., high cosine similarity with each other), they will interact similarly with their respective filter channel slices $F_{j,c}$. LSH allows us to identify such groups of similar input channel slices. In particular, the random projection LSH scheme guarantees \cite{Chen2020SLIDE, Gionis1999Similarity} that the collision probability $p$ for two input channel slices is:
		\begin{equation}
			\label{eq:collision_probability}
			p = \left(1 - \frac{\theta}{\pi} \right)^L \,,
		\end{equation}
		where $\theta \in [0, \pi]$ is the angle between the slices, expressed in radians. Channels with high cosine similarity (and thus, low $\theta$) are therefore highly likely to collide.
		
	\subsection{The \methodname\ Module}
		\label{sec:method:subsec:the_haste_module}
		\begin{figure}[t]
			\centering
			\includegraphics[width=1.0\linewidth]{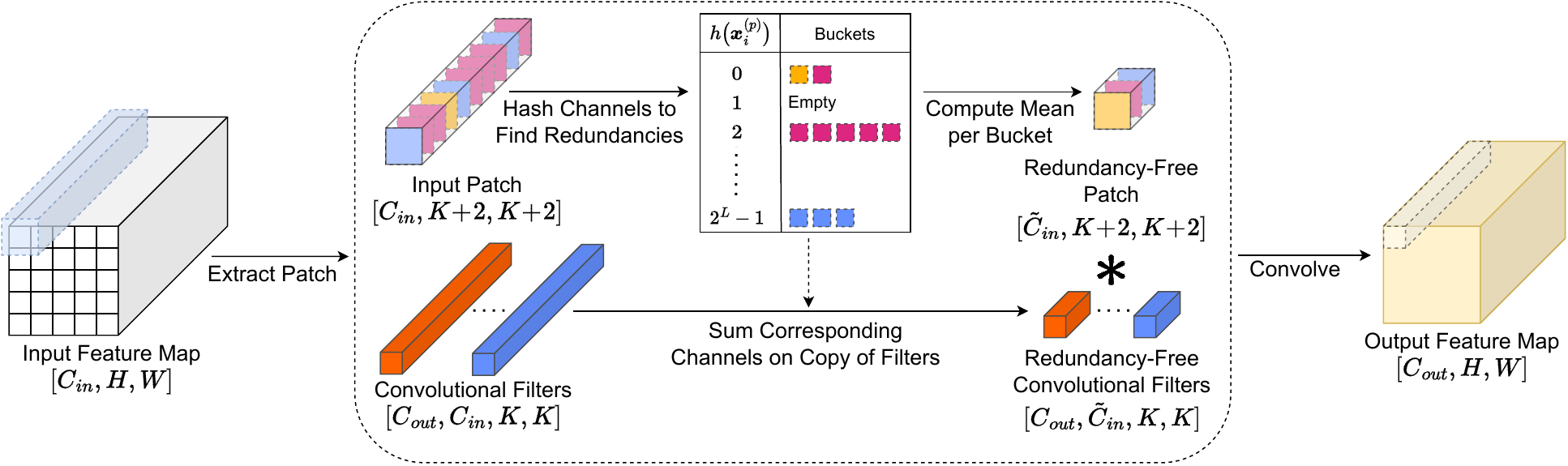}
			\caption{
				Overview of our proposed \methodname\ module. Each patch of the input feature map is processed to find redundant channels. Detected redundancies are then merged together, dynamically reducing the depth of each patch and the convolutional filters. Adapted from \cite{Meiner2025Data}
			}
			\label{fig:method_overview}
		\end{figure}
	
		Suppose a set of input slices $\{X_{c_1}, X_{c_2}, X_{c_3}\}$ at a specific spatial position are grouped into the same hash bucket. We can approximate their contribution to the convolution by a single representative channel slice $\bar{X}_\mathcal{C}$, by taking the average over the channels $\mathcal{C} = \{c_1, c_2, c_3\}$. The contribution of this group in the output $Y_j$ in Equation~\ref{eq:convolution_expressed_as_dot_product} can then be approximated by:
		\begin{equation}
			\label{eq:conv_approximation}
			\sum_{c_i \in \mathcal{C}} F_{j,c_i} \cdot X_{c_i} \approx \sum_{c_i \in \mathcal{C}} F_{j,c_i} \cdot \bar{X}_{\mathcal{C}} = \left(\sum_{c_i \in \mathcal{C}} F_{j,c_i}\right) \cdot \bar{X}_{\mathcal{C}} \,.
		\end{equation}
		This approximation allows computation to be performed in a compressed representation. The terms $\sum F_{j,c_i}$ and $\bar{X}_{\mathcal{C}}$ can be pre-computed once per group. This reduces the number of dot products in the sum, leading to computational savings. Since the grouped filter $\sum F_{j,c_i}$ is applied over many spatial locations with similarly grouped input channel averages, the cost of pre-computation of the filter sum and channel mean is offset by the savings. 
		
		To leverage this approximation inside of a convolutional module, we first process the input feature map $X$. We start by rasterizing the (padded) input features into patches $X_c^{(p)} \in \sR^{(K+2) \times (K+2)}$ for $c \in \{1, \dots, C_{in}\}$, leaving an overlap of two pixels on each side to neighboring patches. Each patch $p$ represents the context window for hashing. We employ patches slightly larger than the kernel size $K$, such that the pre-computed compression is re-used for multiple convolution operations. To identify similar channels for every patch $p$, their representation $X^{(p)}_c$ is flattened into vectors of dimension $(K+2)^2$ and centered by the mean along the channel dimension. By applying the LSH scheme described in Section~\ref{sec:method:sub:lsh} to these vectors, channels with high cosine similarity are grouped in hash buckets. The convolution operation in each group can then be approximated as shown in Equation~\ref{eq:conv_approximation}. The entire procedure is summarized in Algorithm \ref{algo:method}.
		
		In effect, this means that the we dynamically reduce the size of each input context window $X^{(p)}$ by compressing redundant channels. As a result, the number of remaining input channels of a given patch is reduced to $\tilde{C}_{in} \leq C_{in}$, which lets us define a compression ratio $r = 1-(\tilde{C}_{in} / C_{in}) \in [0, 1)$. Note that the compression ratio can vary from patch to patch, as it is based on the redundancy in feature channels at that location. This reduction step is performed on-the-fly for every patch $p$, retaining the original filter weights for the next patch. Since we do not remove entire filters, but only reduce their channel depth, the output feature map retains the same spatial dimension and number of channels as with a regular convolution module. 
		
		In summary, our proposed \methodname\ module addresses the key design considerations for training-free and dynamic model compression:
		\begin{itemize}
			\item \textbf{Computational Overhead.} A main concern is that the method must efficiently detect and compress redundancies at inference time while keeping overhead cost negligible to not undermine FLOPs savings. \methodname\ achieves this through random projections, offering a low computational overhead compared to pairwise distance computation or iterative approaches such as k-nearest neighbors,  as it only requires the computation of $L$ dot products with hyperplane normal vectors.
			\item \textbf{Accuracy.} Without training or fine-tuning, the method must maintain the accuracy of the underlying model. The LSH approach in \methodname\ is designed to only group and merge channels with high cosine similarity (see Equation~\ref{eq:collision_probability}) and thus, high redundancy. It also retains the original input and output dimensions of the model, only compressing repeated features and minimally disrupting information flow.
			\item \textbf{Adjustability.} Real-world deployment scenarios require the method to be adaptable to handle limited energy or compute availability, or constraints related to thermal management. By adjusting the number of hyperplanes $L$ at runtime, we can control the number of hash buckets and thus collisions, directly steering the trade-off between accuracy and computational cost. 
		\end{itemize}
		
		\begin{algorithm}
			\caption{Pseudocode overview of the \methodname\ module. Adapted from \cite{Meiner2025Data}}
			\label{algo:method}
			\begin{algorithmic}[1]
				\Require Feature map $X \in \sR^{C_{in} \times H \times W}$, Filters $F \in \sR^{C_{out} \times C_{in} \times K \times K}$ 
				\Ensure $Y \in \sR^{C_{out} \times H \times W}$ \\
				\textbf{Initialize:} $h: \sR^{(K+2)^2} \to \{0, \dots, 2^L-1\}$ 
				\For{every patch $p$} \Comment{Iterate over every patch in the spatial grid.}
				\State HashCodes = [ ]
				\For{$c = 1, \dots, C_{in}$}
				\State $x_c^{(p)} = $ Center(Flatten($X^{(p)}_c$)) 
				\State HashCodes.Append($h(x_c^{(p)})$) \Comment{Hash the flattened input patch.}
				\EndFor
				\State $\bar{X}^{(p)}$ = MergeInput($X^{(p)}$, HashCodes) \Comment{Take mean of redundant channels.}
				\State $\tilde{F}$ = MergeFilters($F$, HashCodes) \Comment{Take sum of redundant channels.}
				\State $Y^{(p)}$ = $\bar{X}^{(p)}$ * $\tilde{F}$ \Comment{Approximate convolution as in Equation~\ref{eq:conv_approximation}.}
				\EndFor
				\State \Return $Y$
			\end{algorithmic}
		\end{algorithm}
	
	\subsection{Efficient Hashing with Sparse Hyperplanes}
		Locality-sensitive hashing significantly reduces computational overhead compared to exact nearest neighbor searches by using random hyperplanes to generate binary code identifiers. However, using a typical LSH implementation in a \methodname\ module still incurs some overhead computational cost. Specifically, it requires $L \cdot d$ multiplications and $L \cdot (d-1)$ additions per input, where $d$ is the dimensionality of the input vectors.
		
		To address this remaining overhead, we employ sparse random projections as proposed by \citet{Achlioptas2003Database} and \citet{Li2006Very}. Instead of generating hyperplane normal vectors $v_l$ from standard normal distributions, we construct very sparse vectors $\tilde{v}_l$ whose entries are constrained to the set $\{1, 0, -1\}$. We control their degree of sparsity using a hyperparameter $s \in (0,1)$, representing the expected fraction of zero entries. Non-zero entries are randomly assigned values of $+1$ or $-1$ with equal probability.
		
		Using sparse hyperplanes lowers the computational requirements further: dot product calculations now consist solely of additions, eliminating multiplications entirely. Specifically, each hyperplane computation requires only $L\cdot (d(1-s)-1)$ additions. This enhances runtime efficiency, as computationally expensive multiplications are replaced by cheaper additions.
		
		Our method introduces two hyperparameters: the number of hyperplanes $L$ and their sparsity $s$. By adjusting $L$, we can directly steer the trade-off between the degree of compression and model accuracy. Therefore, practitioners can adjust this parameter based on the application's needs in a flexible manner, targeting either high computational efficiency or accuracy retention. 
		On the other hand, the sparsity parameter $s$ does not need intensive tuning. It can typically be set consistently across all models trained on a specific dataset. The theoretical insights provided by \citet{Achlioptas2003Database} and \citet{Li2006Very} provide good starting choices for $s$. We further detail these hyperparameter choices in Section~\ref{sec:experiments:subsec:experiment_settings} and analyze the impact of $s$ in Section~\ref{sec:experiments:subsec:ablations:subsubsec:hyperplane_sparsity}.
		
		\begin{figure*}[t]
			\centering
			\begin{subfigure}{1\textwidth}
				\includegraphics[width=1\linewidth]{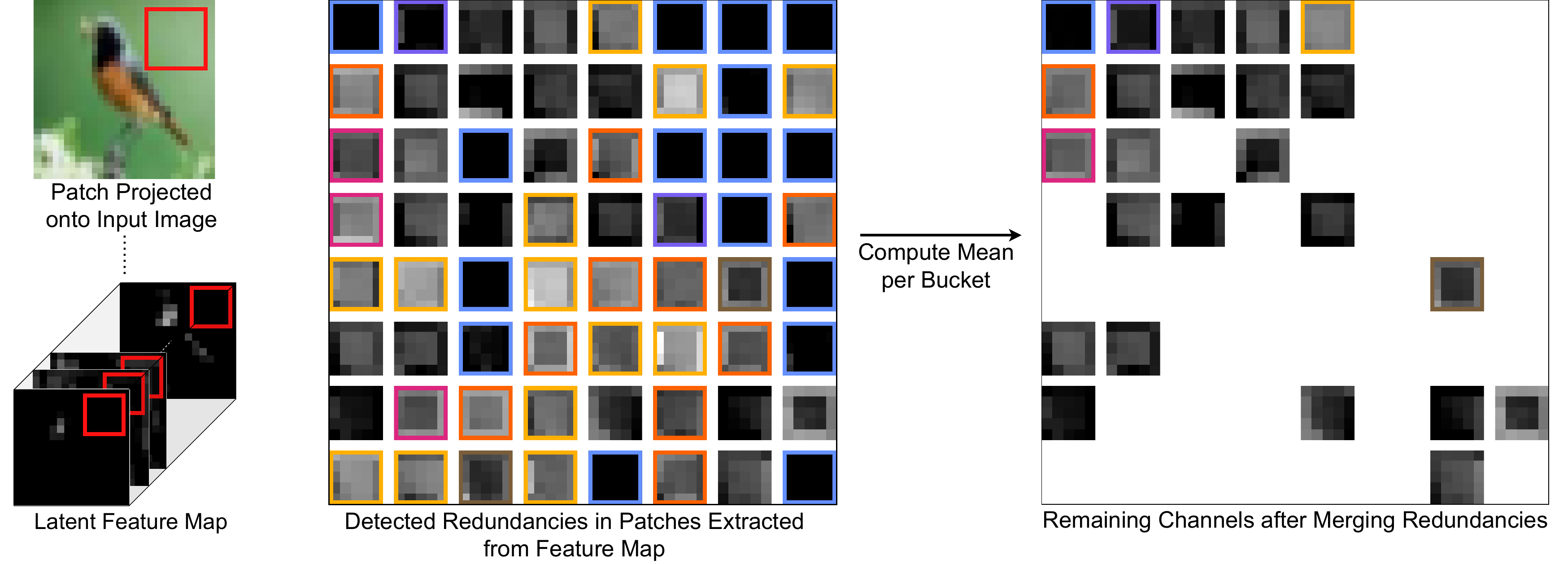}
				\caption{A patch from a low-complexity background region. Our method detects high redundancy, allowing us to merge many channels. Here, the input channel dimension is reduced from 64 to 24, resulting in a compression ratio of $62.50\%$}
			\end{subfigure}
			\begin{subfigure}{1\textwidth}
				\includegraphics[width=1\linewidth]{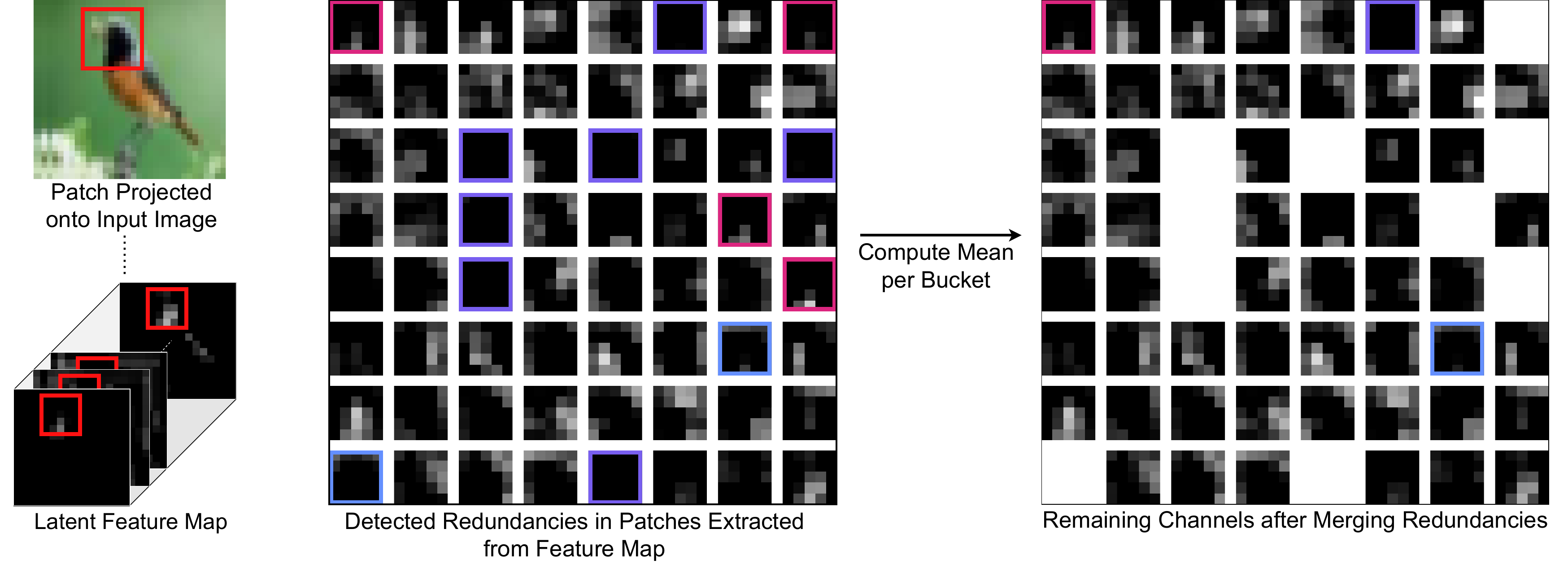}
				\caption{A patch containing more complex textures. Fewer redundancies are detected, as the features are more varied. Here, the input channel dimension is reduced from 64 to 54, resulting in a compression ratio of $15.63\%$}
			\end{subfigure}
			\caption{
				Visualization of the dynamic, input-dependent channel compression performed by the \methodname\ module in a ResNet18 model on CIFAR-10. For two different patches from the same input image, our method identifies varying degrees of redundancy in the latent features based on LSH. Channels sharing the same outline color fall into the same hash bucket and are merged by averaging. Patches with no matching hash code (and therefore, no outline) are left unchanged. Adapted from \cite{Meiner2025Data}
			}
			\label{fig:feature_map_visualization}
		\end{figure*}

\section{Experiments}
	\label{sec:experiments}

	In this section, we evaluate the effectiveness of our plug-and-play compression approach on standard convolutional neural network architectures, focusing on both the reduction in floating point operations and the preservation of model accuracy. We begin by detailing the experimental setup in Section~\ref{sec:experiments:subsec:experiment_settings}, followed by a verification of the design principles of \methodname\ in Section~\ref{sec:experiments:subsec:analysis_of_haste_module}. Next, we present our results on the CIFAR-10 \cite{Krizhevsky2009Learning} and ImageNet ILSVRC 2012 (ImageNet-1K) \cite{Russakovsky2015ImageNet} datasets in Sections~\ref{sec:experiments:subsec:results_on_cifar10} and~\ref{sec:experiments:subsec:results_on_imagenet}, respectively, where we also compare our method to related approaches. We also examine the adjustability of our method with respect to the hyperparameter $L$ as well as its scaling behavior with larger model architectures. Furthermore, we extend on \cite{Meiner2025Data} by conducting ablations on the context patch size, the hyperplane sparsity setting $s$ as well as the choice of starting layer for \methodname\ application in Section~\ref{sec:experiments:subsec:ablation_studies}. We close our analysis by evaluating the out-of-the-box performance of \methodname\ on lightweight CNN architectures in Section~\ref{sec:experiments:subsec:performance_on_lightweight_architectures}.
	
	\subsection{Experiment Settings}
		\label{sec:experiments:subsec:experiment_settings}
		
		\begin{table}
			\centering
			\caption{
				Overview of related pruning approaches. While other methods require either fine-tuning or a specialized training procedure to achieve notable FLOPs reduction, our method is completely training-free and uniquely offers adjustable compression ratios at runtime. Adapted from \cite{Meiner2025Data}
			}
			\label{tab:other_approaches_overview}
			\begin{tabular}{lcccc}
				\toprule
				Method & Dynamic Inference & No Training & No Fine-Tuning & Adaptable at Runtime\\
				\midrule
				DGNet \cite{Li2021Dynamic} & \cmark & \xmark & \cmark & \xmark \\
				DMCP \cite{Xu2021Efficient} & \cmark & \xmark & \cmark & \xmark \\
				DynConv \cite{Verelst2020Dynamic} & \cmark & \xmark & \cmark & \xmark \\
				FBS \cite{Gao2019Dynamic} & \cmark & \xmark & \cmark & \xmark \\
				FPGM \cite{He2019Filter} & \xmark & \xmark & \cmark & \xmark \\
				FTWT \cite{Elkerdawy2022Fire} & \cmark & \xmark & \cmark & \xmark \\
				LCCN \cite{Dong2017More} & \cmark & \xmark & \cmark & \xmark \\
				PFEC \cite{Li2017Pruning} & \xmark & \cmark & \xmark & \xmark \\
				SSL \cite{Wen2016Learning} & \xmark & \cmark & \xmark & \xmark \\
				\midrule
				\textbf{\methodname\ (ours)} & \cmark & \cmark & \cmark & \cmark \\
				\bottomrule 
			\end{tabular}
		\end{table}
		
		For our experiments, we use publicly available pre-trained models \cite{Phan2021huyvnphan/PyTorch_CIFAR10, Paszke2019PyTorch}. In these models, we replace standard, non-strided convolutions with our \methodname\ module. In ResNet architectures \cite{He2016Deep}, we exclude downsampling layers from this process.
		
		We adapt the hyperplane sparsity $s$ and the starting layer for the use of \methodname\ based on the dataset. For CIFAR-10 \cite{Krizhevsky2009Learning}, we use a high sparsity setting of $s=2/3$ as suggested by \citet{Achlioptas2003Database} and apply \methodname\ early, from the first convolutional layer in VGG models \cite{Simonyan2015Very} and the first residual block after max pooling in ResNet architectures \cite{He2016Deep}. For the more complex ImageNet \cite{Russakovsky2015ImageNet} dataset, we lower the sparsity to $s=1/2$ and apply \methodname\ later in the models, starting from the third convolution in VGG and the second layer in ResNets and WideResNets \cite{Zagoruyko2016Wide}. This strategy accounts for the lower redundancy in the latent features of the initial layers in ImageNet models. We provide an ablation of these settings in Section~\ref{sec:experiments:subsec:ablation_studies}.
		
		Crucially, our method requires no training or fine-tuning, so models are evaluated on a single NVIDIA Tesla T4 GPU immediately after inserting the \methodname\ modules. We report the mean top-1 accuracy and FLOPs reduction over three random seeds, including standard deviations. To evaluate our approach, we use the CIFAR-10 test set and the ILSVRC 2012 validation set for ImageNet. Additionally, we provide latency estimates derived from measurements on an Intel i7-11850H CPU (see Table \ref{tab:latency} and Section~\ref{sec:experiments:subsec:results_on_cifar10:subsubsection:latency_and_memory_analysis}).
		
		As, to our knowledge, \methodname\ is the first entirely training-free CNN compression technique adjustable at runtime, direct comparisons are not possible. We therefore test it against related, state-of-the-art channel pruning and dynamic gating methods that require dedicated training or tuning stages (see Table~\ref{tab:other_approaches_overview} for an overview).
	
	\subsection{Analysis of the \methodname\ Module}
	\label{sec:experiments:subsec:analysis_of_haste_module}
	
		To isolate the contributions of our method's key components, we conduct experiments comparing \methodname\ against several training-free baseline configurations. These baselines are constructed by varying three core design choices:
		\begin{enumerate}
			\item \textbf{Compression Scope.} Compression is either applied \textit{globally} to entire input channels or on a \textit{patch-wise} basis, as proposed in our method (see Section~\ref{sec:method:subsec:the_haste_module}).
			\item \textbf{Compression Criterion.} The decision of which channels to compress is based on either the channel's $L^1$ norm \cite{Li2017Pruning} or our proposed \textit{LSH} scheme for detecting redundancies (see Section~\ref{sec:method:subsec:finding_redundancies}). 
			\item \textbf{Compression Operation.} Selected channels are either completely \textit{removed} or grouped and \textit{merged} into a single representation.
		\end{enumerate}
		
		Our \methodname\ module uniquely combines the patch-wise scope, an LSH-based compression criterion and channel merging. We create four baseline variants using the $L^1$ norm criterion and systematically explore the other two settings: global channel removal, global channel merging, patch-wise removal and patch-wise merging. For a fair comparison, all methods are configured to achieve a similar compression ratio per model on the CIFAR-10 dataset without any fine-tuning.
		
		The results presented in Table~\ref{tab:method_component_ablation} demonstrate the benefits of our approach. While baselines using patch-wise operations improve over global pruning or merging, only the full \methodname\ configuration, which uses LSH to detect redundant channels, consistently maintains near-baseline accuracy. The naive $L^1$ norm-based methods, even with patch-wise operation, result in a significant degradation in performance.

		\begin{table}[t]
			\centering
			\caption{Comparing our full \methodname\ method against four training-free baseline variants on CIFAR-10. The baselines are constructed by varying the \textit{Compression Scope} (Global vs. Patch-wise), the \textit{Compression Criterion} ($L^1$ norm vs. LSH), and the \textit{Compression Operation} (Remove vs. Merge). \methodname\ combines the patch-wise scope, LSH criterion, and merge operation. For each model, we report top-1 accuracy (\%) and accuracy drop ($\Delta$) after compression. We highlight the best accuracy score after compression and lowest accuracy drop ($\Delta$) in \textbf{bold}. Adapted from \cite{Meiner2025Data}}
			\label{tab:method_component_ablation}
			\begin{tabular*}{\textwidth}{@{} ccccc @{}}
				\toprule
				Method  & ResNet18 & ResNet34 & VGG11-BN & VGG19-BN \\
				(Scope + Criterion + Op.) & (93.07) & (93.34) & (92.39) & (93.95) \\
				\midrule
				Compression Target & 40\% & 50\% & 40\% & 40\% \\
				\midrule
				Global + $L^1$ + Remove & 71.07 (-22.00) & 48.42 (-44.92) & 41.77 (-50.62) & 34.89 (-59.06) \\
				Global + $L^1$ + Merge & 65.31 (-27.76) & 40.52 (-52.82) & 73.87 (-18.52) & 42.23 (-51.72) \\
				Patch + $L^1$ + Remove & 88.70 (-4.37) & 80.04 (-13.30) & 65.94 (-25.45) & 65.84 (-28.11) \\
				Patch + $L^1$ + Merge & 86.53 (-6.54) & 72.10 (-21.24) & 87.39 (-5.00) & 82.51 (-11.44) \\
				\textbf{Ours (\methodname)} & \textbf{91.18 (-1.89)} & \textbf{90.45 (-2.89)} & \textbf{89.36 (-3.03)} & \textbf{91.19 (-2.76)} \\
				\bottomrule
			\end{tabular*}
		\end{table}
	
	\subsection{Results on CIFAR-10}
	\label{sec:experiments:subsec:results_on_cifar10}
	
\begin{figure}[t]
			\centering
			\begin{subfigure}{0.45\textwidth}
				\centering
				\includegraphics[width=0.95\textwidth]{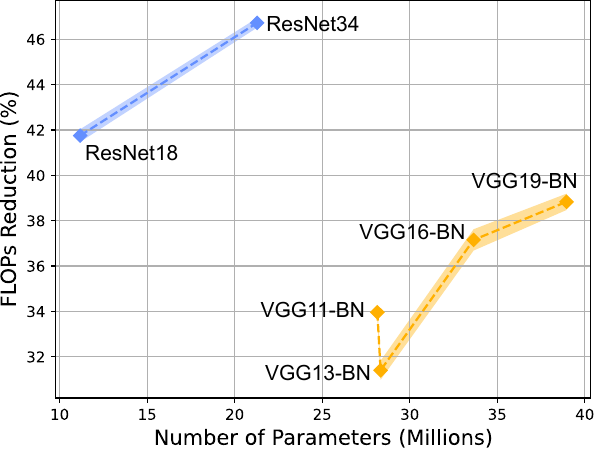}
				\caption{Overview of CIFAR-10 results}
				\label{fig:results_cifar10:subfig:cifar10_overview}
			\end{subfigure}
			\hfill
			\begin{subfigure}{0.45\textwidth}
				\centering
				\includegraphics[width=0.95\textwidth]{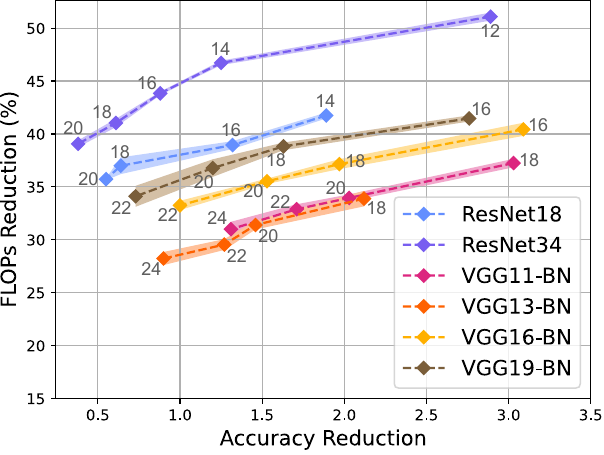}
				\caption{Influence of hyperparameter $L$}
				\label{fig:results_cifar10:subfig:cifar10_pareto}
			\end{subfigure}
			\caption{Results of our method on the CIFAR-10 dataset. (a) shows the achieved FLOPs reduction
				for all tested models, using $L = 14$ for ResNets and $L = 20$ for VGG-BN models. (b) depicts the
				influence of the chosen number of hyperplanes $L$ (shown in \textcolor{gray}{gray}) on compression rates and accuracy. Reproduced from \cite{Meiner2025Data}}
			\label{fig:results_cifar10}
		\end{figure}

		We evaluate \methodname\ on a range of ResNet and VGG-BN architectures on the CIFAR-10 dataset. Our training-free method demonstrates the ability to deliver significant computational savings while preserving high model accuracy, without any fine-tuning. For instance, on ResNet34, \methodname\ reduces FLOPs by 46.72\% with only a 1.25 percentage point drop in top-1 accuracy. The results are visualized in Figure~\ref{fig:results_cifar10}.
		
		Table \ref{tab:results_c10} presents a comparative analysis against state-of-the-art pruning methods for CNN architectures. Despite its training-free nature, \methodname\ achieves performance competitive with techniques that require data-dependent training or fine-tuning after the compression step. While the accuracy of baseline models differs due to different pre-trained checkpoints, our results for ResNet18 and VGG19-BN are comparable with methods that require training, such as DMCP \cite{Xu2021Efficient} and SSL \cite{Wen2016Learning}. For the ResNet18 model, we provide visualizations of the resulting channel clusters in Figure~\ref{fig:feature_map_visualization}.
		
		\begin{table}[t]
			\centering
			\caption{Selected results on CIFAR-10. "FLOPs Red." denotes the percentage decrease of FLOPs after pruning compared to the base model. Reproduced from \cite{Meiner2025Data}}
			\label{tab:results_c10}
				\begin{tabular*}{\linewidth}{@{\extracolsep{\fill}}ccccccc}
					\toprule
					\multirow[c]{2.5}{*}{Model} & \multirow[c]{2.5}{*}{Method} & \multicolumn{3}{c}{Top-1 Accuracy (\%)} & \multirow[c]{2.5}{*}{\makecell{FLOPs Red.\ \\(\%)}} & \multirow[c]{2.5}{*}{\makecell{Training-\\Free}}\\
					\cmidrule(lr){3-5}
					& & Baseline & Pruned & $\Delta$ & & \\
					\midrule
					\multirow[c]{5.5}{*}{\rotatebox{90}{ResNet18}} & PFEC$^*$ & 91.38 & 89.63 & 1.75 & 11.71 & \xmark \\
					& SSL$^*$ & 92.79 & 92.45 & 0.34 & 14.69 & \xmark \\ 
					& DMCP & 92.87 & 92.61 & 0.26 & 35.27 & \xmark  \\ 
					\cmidrule{2-7}
					& \textbf{Ours} ($L=14$) & 93.07 & 91.18 $(\pm 0.38)$ & 1.89 & 41.75 $(\pm 0.28)$ & \cmark \\
					& \textbf{Ours} ($L=20$) & 93.07 & 92.52 $(\pm 0.10)$ & 0.55 & 35.73 $(\pm 0.09)$ & \cmark \\
					\midrule
					\multirow[c]{6.5}{*}{\rotatebox{90}{VGG16-BN}} & PFEC$^*$ & 91.85 & 91.29 & 0.56 & 13.89 & \xmark \\
					& SSL$^*$ & 92.09 & 91.80 & 0.29 & 17.76 & \xmark\\
					& DMCP & 92.21 & 92.04 & 0.17 & 25.05 & \xmark \\ 
					& FTWT & 93.82 & 93.73 & 0.09 & 44.00 & \xmark \\
					\cmidrule{2-7}
					& \textbf{Ours} ($L=18$) & 94.00 & 92.03 $(\pm 0.21)$ & 1.97 & 37.15 $(\pm 0.47)$ & \cmark \\
					& \textbf{Ours} ($L=22$) & 94.00 & 93.00 $(\pm 0.12)$ & 1.00 & 33.25 $(\pm 0.44)$ & \cmark \\
					\midrule
					\multirow[c]{5.5}{*}{\rotatebox{90}{VGG19-BN}} & PFEC$^*$ & 92.11 & 91.78 & 0.33 & 16.55 & \xmark \\
					& SSL$^*$ & 92.02 & 91.60 & 0.42 & 30.68 & \xmark \\
					& DMCP  & 92.19 & 91.94 & 0.25 & 34.14 & \xmark \\ 
					\cmidrule{2-7}
					& \textbf{Ours} ($L=18$) & 93.95 & 92.32 $(\pm 0.35)$ & 1.63 & 38.83 $(\pm 0.36)$ & \cmark \\
					& \textbf{Ours} ($L=22$) & 93.95 & 93.22 $(\pm 0.14)$ & 0.73 & 34.11 $(\pm 0.99)$ & \cmark \\
					\bottomrule 
				\end{tabular*}
				\begin{tablenotes}\footnotesize
					\item[*] Results taken from \citet{Xu2021Efficient}.
				\end{tablenotes}
		\end{table}
		
		\subsubsection{Adjustable Accuracy-FLOPs Trade-off}
			A key feature of \methodname\ is the ability to adjust the trade-off between compression and retained accuracy at inference time. By varying the hyperparameter $L$, the number of hyperplanes used in our LSH scheme, we can navigate the Pareto front of performance, as depicted in Figure \ref{fig:results_cifar10:subfig:cifar10_pareto}. A larger $L$ leads to more conservative compression, preserving more accuracy at the cost of lower FLOPs savings. For example, on ResNet34, the accuracy drop can be tuned from 2.89 points (for 51.09\% FLOPs reduction with $L=12$) to only 0.38 points (for 39.07\% FLOPs reduction with $L=20$).
		
		\subsubsection{Latency and Memory Analysis}
			\label{sec:experiments:subsec:results_on_cifar10:subsubsection:latency_and_memory_analysis}
			Beyond theoretical FLOPs, we assess the practical performance of \methodname\ in terms of inference latency and memory usage.
			
			\begin{table}[b]
				\centering
				\caption{Latency estimates for \methodname\ on CIFAR-10 and ImageNet. We report baseline and compressed latencies in milliseconds (ms). The numbers in brackets denote the speedup compared to the baseline. The realistic setting assumes hardware support for patch-wise operations. The theoretical speedup is derived from the achieved FLOPs reduction. Adapted from \cite{Meiner2025Data}}
				\label{tab:latency}
				\begin{tabular*}{\linewidth}{ ccccc }
					\toprule
					Dataset & \multicolumn{2}{c}{CIFAR-10} & \multicolumn{2}{c}{ImageNet} \\
					\cmidrule(r){1-1} \cmidrule(lr){2-3} \cmidrule(l){4-5}
					\multirow[c]{2}{*}{Model} & ResNet18 & ResNet34 & ResNet34 & VGG19-BN \\
					& ($L=14$) & ($L=14$) & ($L=16$) & ($L=20$) \\
					\midrule
					Baseline & $8.73\,$ms & $15.54\,$ms & $103.50\,$ms & $476.96\,$ms \\
					Realistic & $5.88\,$ms ($1.48\times$) & $10.60\,$ms ($1.47\times$) & $84.56\,$ms ($1.22\times$) & $371.59\,$ms ($1.28\times$) \\
					Theoretical & $5.09\,$ms ($1.72\times$) & $8.28\,$ ms ($1.88\times$) & $80.06\,$ms ($1.29\times$) & $329.91\,$ ms ($1.45\times$) \\
					\bottomrule
				\end{tabular*}
			\end{table}
		
			\bmhead{Latency} Directly measuring the latency of \methodname\ in standard deep learning frameworks is misleading, as they are optimized for dense, static computations without native support for the dynamic, conditional operations \cite{Belcak2023Exponentially} found in our method. Therefore, we choose to provide estimates for the potential speedup, which we report in Table~\ref{tab:latency}. 
			
			The \textit{realistic} setting denotes an estimate derived from a latency-per-FLOP measurement of the baseline model, and extrapolating it to the FLOPs of each individual component (hashing, merging, reduced convolution). This assumes the existence of optimized hardware or software than can efficiently execute dynamic, patch-wise operations, and accounts for the overhead of our hashing scheme. The \textit{theoretical} scenario provides a simpler upper-bound estimate by reducing the baseline model's latency proportionately to the total measured FLOPs reduction.
			
			As shown in Table \ref{tab:latency}, we calculate that our method accelerates ResNet18 and ResNet34 by a factor of approximately 1.5$\times$, assuming realistic hardware support for the patch-wise operations on a CPU. This demonstrates a practical benefit that closely follows the theoretical speedup derived from FLOPs reduction.
			
			\begin{figure}
				\centering
				\includegraphics[width=1\linewidth]{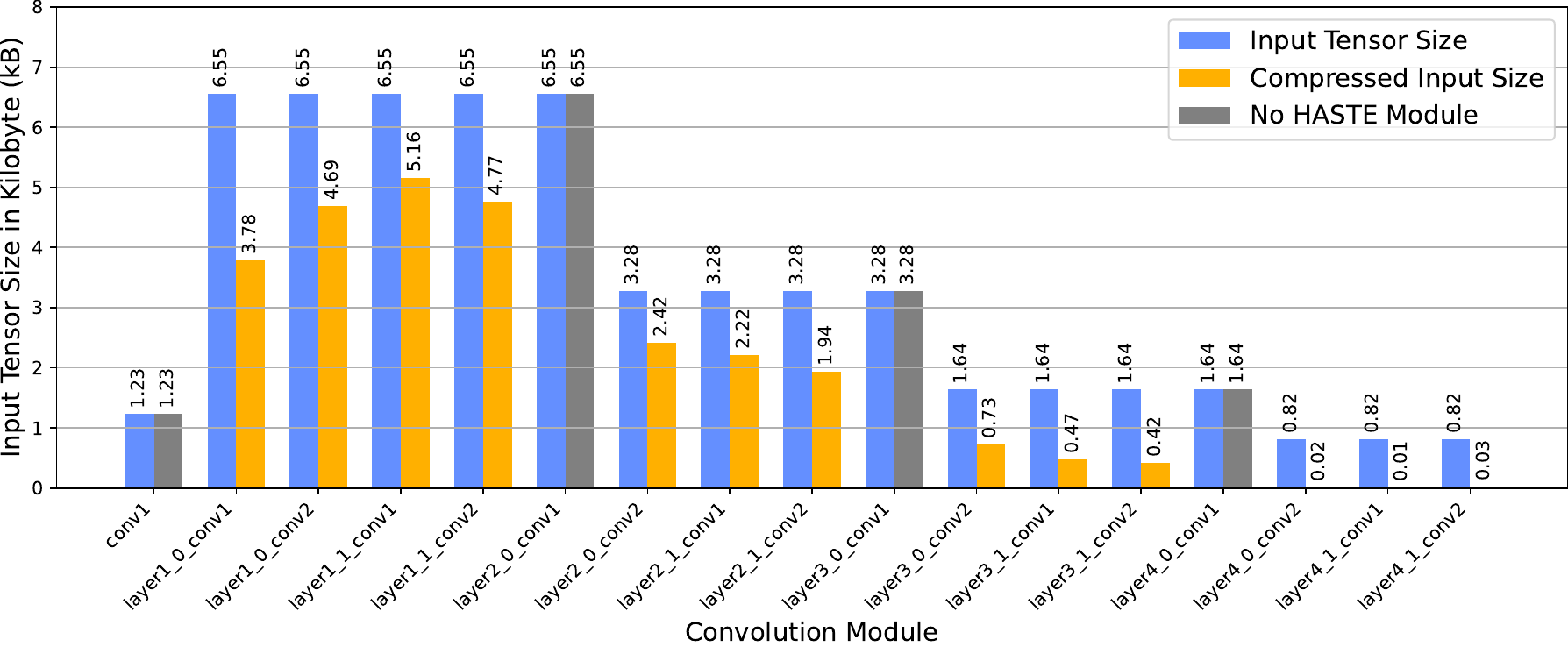}
				\caption{Memory requirements for input tensors in ResNet18 ($L=14$) on CIFAR-10}
				\label{fig:memory_cifar10}
			\end{figure}
			
			\bmhead{Memory} Regarding memory, \methodname\ offers an advantage in reducing activation memory while incurring negligible parameter overhead. While the base model weights are retained (unlike in static pruning), our method compresses the intermediate feature maps before the convolution. This dynamic compression is crucial for reducing memory bus traffic, a key bottleneck on embedded hardware \cite{Vogel2019Guaranteed}. The average compression ratio achieved directly translates to an equivalent reduction in memory required for these latent tensors. Figure \ref{fig:memory_cifar10} visualizes this effect, showing that input tensor sizes are drastically reduced, especially in deeper layers.
			
			Furthermore, we note that the parameter storage overhead of the \methodname\ modules is minimal. The additional parameters consist only of the sparse, ternary-valued normal vectors that define the hyperplanes used for LSH. The relative memory cost of a \methodname\ module compared to its corresponding convolution is:
			\begin{equation}
				\frac{\text{Mem(HASTE)}}{\text{Mem(Conv)}} = 1 + \frac{3 \cdot L \cdot (K+2)^2}{32 \cdot C_\text{out} \cdot 		C_\text{in} \cdot K^2}.
			\end{equation}
			For a typical configuration (ResNet18, $L=20$), this results in an average parameter storage memory overhead of just 0.04\% per module, making our approach highly suitable for memory-constrained environments.
	
	\subsection{Results on ImageNet}
		\label{sec:experiments:subsec:results_on_imagenet}
	
		We benchmark \methodname\ on the large-scale ImageNet dataset using a variety of ResNet, WideResNet, and VGG-BN models. The increased complexity of ImageNet, featuring 100$\times$ more classes and approximately 26$\times$ more images than CIFAR-10, results in latent features with less redundancy. Consequently, while our method still provides notable computational savings, the achievable compression rates are more modest compared to those on CIFAR-10. This highlights a fundamental trade-off: as a model's learned representations become richer and less redundant, training-free compression becomes inherently more challenging.
	
		Table \ref{tab:results_imagenet} compares \methodname\ against several prominent data-dependent pruning and dynamic execution methods. While approaches that leverage training data for extensive tuning achieve higher FLOPs reduction, our approach provides a strong, data-free baseline. For instance, on ResNet34, \methodname\ reduces FLOPs by 18.69\% with only a 1.25 percentage point accuracy drop, without requiring access to a single data sample.
		
		\begin{figure}[t]
			\centering
			\begin{subfigure}{0.45\textwidth}
				\centering
				\includegraphics[width=0.95\textwidth]{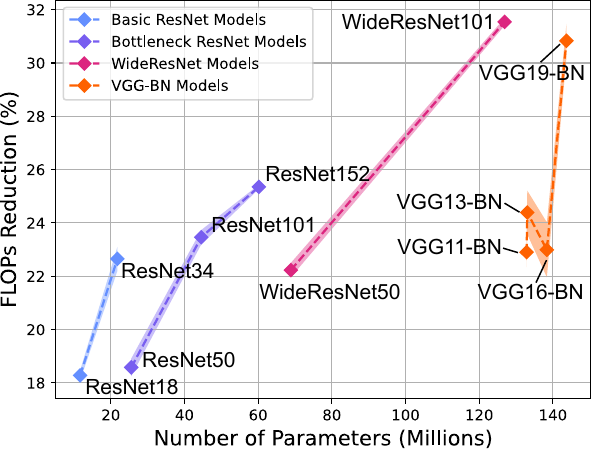}
				\caption{Overview of ImageNet experiments}
				\label{fig:imagenet_results:subfig:imagenet_overview}
			\end{subfigure}
			\hfill
			\begin{subfigure}{0.45\textwidth}
				\centering
				\includegraphics[width=0.95\textwidth]{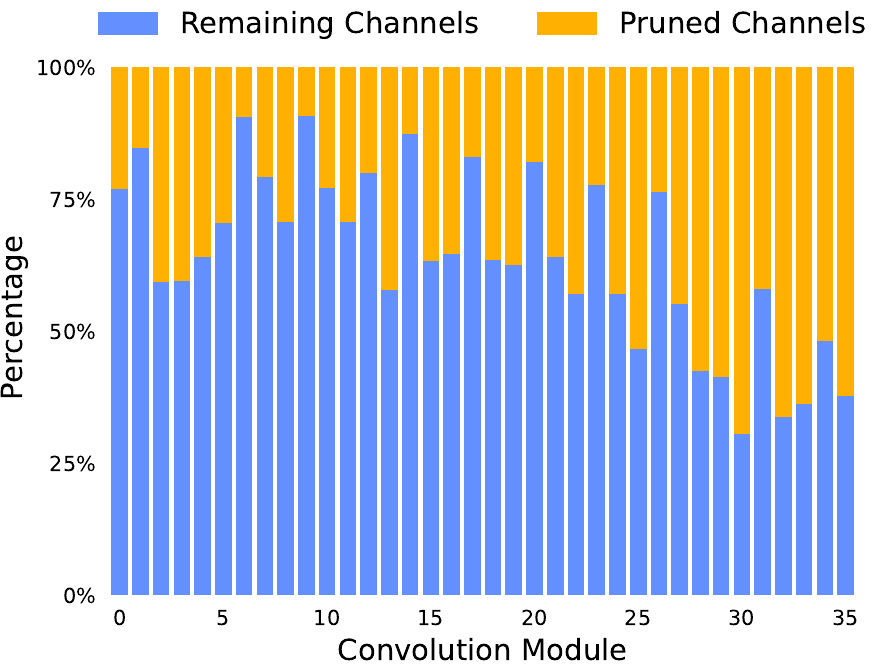}
				\caption{Distribution of compressed channels}
				\label{fig:imagenet_results:subfig:resnet50_gating}
			\end{subfigure}
			\caption{Visualization of results on the ImageNet dataset. (a) depicts the trade-off between FLOPs reduction and number of parameters for all tested architectures. Results are shown with $L=16$ for basic ResNet models, $L=28$ for bottleneck ResNets, $L=32$ for WideResNets, and $L=20$ for VGG-BN models. (b) shows the achieved compression rate per convolution module in a ResNet50, starting from the second bottleneck layer. Reproduced from \cite{Meiner2025Data}}
			\label{fig:imagenet_results}
		\end{figure}
	
		\begin{table}[t]
			\centering
			\caption{Selected results on ImageNet. "FLOPs Red." denotes the percentage reduction of FLOPs after pruning compared to the baseline. Reproduced from \cite{Meiner2025Data}}
			\label{tab:results_imagenet}
			\begin{tabular*}{\linewidth}{@{\extracolsep{\fill}} ccccccc}
				\toprule
				\multirow[c]{2.5}{*}{Model} & \multirow[c]{2.5}{*}{Method} & \multicolumn{3}{c}{Top-1 Accuracy (\%)} & \multirow[c]{2.5}{*}{\makecell{FLOPs Red.\ \\(\%)}} & \multirow[c]{2.5}{*}{\makecell{Training-\\Free}}\\
				\cmidrule(lr){3-5}
				& & Baseline & Pruned & $\Delta$ & & \\
				\midrule
				\multirow[c]{7.5}{*}{\rotatebox{90}{ResNet18}} & LCCN & 69.98 & 66.33 & 3.65 & 34.60 & \xmark \\
				& DynConv$^*$ & 69.76 & 66.97 & 2.79 & 41.50 & \xmark  \\ 
				& FPGM & 70.28 & 68.34 & 1.94 & 41.80 & \xmark  \\
				& FBS & 70.71 & 68.17 & 2.54 & 49.49 & \xmark \\
				& FTWT & 69.76 & 67.49 & 2.27 & 51.56 & \xmark \\
				\cmidrule{2-7}
				& \textbf{Ours} ($L=16$) & 69.76 & 66.97 $(\pm 0.21)$ & 2.79 & 18.28 $(\pm 0.19)$ & \cmark \\
				& \textbf{Ours} ($L=20$) & 69.76 & 68.64 $(\pm 0.56)$ & 1.12 & 15.10 $(\pm 0.18)$ & \cmark \\
				\midrule
				\multirow[c]{7.5}{*}{\rotatebox{90}{ResNet34}} & PFEC & 73.23 & 72.09 & 1.14 & 24.20 & \xmark  \\
				& LCCN & 73.42 & 72.99 & 0.43 & 24.80 & \xmark \\
				& FPGM & 73.92 & 72.54 & 1.38 & 41.10 & \xmark \\
				& FTWT & 73.30 & 72.17 & 1.13 & 47.42 & \xmark \\
				& DGNet & 73.31 & 71.95 & 1.36 & 67.20 & \xmark \\
				\cmidrule{2-7}
				& \textbf{Ours} ($L=16$) & 73.31 & 70.31 $(\pm 0.07)$ & 3.00 & 22.65 $(\pm 0.45)$ & \cmark \\
				& \textbf{Ours} ($L=20$) & 73.31 & 72.06 $(\pm 0.05)$ & 1.25 & 18.69 $(\pm 0.30)$ & \cmark \\
				\midrule
				\multirow[c]{4.5}{*}{\rotatebox{90}{ResNet50}} & FPGM & 76.15 & 74.83 & 1.32 & 53.50 & \xmark \\
				& DGNet & 76.13 & 75.12 & 1.01 & 67.90 & \xmark \\
				\cmidrule{2-7}
				& \textbf{Ours} ($L=28$) & 76.13 & 73.04 $(\pm 0.07)$ & 3.09 & 18.58 $(\pm 0.33)$ & \cmark \\
				& \textbf{Ours} ($L=36$) & 76.13 & 74.77 $(\pm 0.10)$ & 1.36 & 15.68 $(\pm 0.16)$ & \cmark \\
				\bottomrule 
			\end{tabular*}
			\begin{tablenotes}\footnotesize
				\item[*] Results taken from \cite{Li2021Dynamic}.
			\end{tablenotes}
		\end{table}
		
		\subsubsection{Scaling with Model Size and Width}
		
			A key finding is that the effectiveness of \methodname\ scales positively with model size and width, as shown in Figure~\ref{fig:imagenet_results:subfig:imagenet_overview}. Larger models tend to exhibit greater redundancy, which our method successfully exploits. This trend is particularly noticeable in the WideResNet family. For example, we achieve up to a 31.54\% FLOPs reduction on WideResNet101. This suggests that \methodname\ is especially well-suited for compressing highly overparameterized models.
			
			Analyzing the per-layer compression rates in a ResNet50, as depicted in Figure~\ref{fig:imagenet_results:subfig:resnet50_gating}, reveals that our method dynamically adapts to varying degrees of redundancy in latent features. We observe more aggressive compression in deeper layers, where feature representations tend to capture more high-level global information \cite{Zeiler2014Visualizing}.
		
		\subsubsection{Impact of Pointwise Convolutions and Latency}
			\label{sec:experiments:subsec:results_on_imagenet:subsubsec:impact_of_pointwise_convolutions}
			Our analysis also reveals the impact of network architecture on performance. A noticeable dip in FLOPs reduction occurs when moving from standard residual blocks (ResNet34) to bottleneck blocks (ResNet50).
			This is not necessarily because 1$\times$1 convolutions are harder to compress. In fact, Figure~\ref{fig:imagenet_results:subfig:resnet50_gating} shows they can be compressed effectively. Instead, we mainly relate this issue to the fact that our LSH scheme is proportionately more expensive for pointwise convolutions. While the number of FLOPs required to perform a 1$\times$1 convolution is 9$\times$ lower than for 3$\times$3 convolutions of the same channel dimensions, the hashing cost is only 2.8$\times$ lower, making LSH relatively costly. Thereby, the fixed computational overhead of hashing and merging operations constitutes a larger portion of the total cost, lowering the percentage of FLOPs saved.
			
			Despite this, the theoretical savings still hint at practical gains. As detailed in Table~\ref{tab:latency}, our method has potential to deliver tangible inference speedup on a CPU, accelerating a ResNet34 by a factor of 1.22$\times$ and a VGG19-BN by 1.28$\times$.
	
	\subsection{Ablation Studies}
		\label{sec:experiments:subsec:ablation_studies}
		
		\subsubsection{Patch Size}
			The patch size is a core design choice in \methodname, not a tunable hyperparameter. For a given convolution module with kernel size $K \times K$, we set the patch size to $(K+2) \times (K+2)$. This is the minimal size that allows the filter kernel to perform nine convolution operations per patch, reusing the pre-computed compressed input and filter representations (see Section~\ref{sec:method:subsec:the_haste_module}). This localized context also increases channel-wise redundancies, which our LSH-based merging is designed to leverage.
		
\begin{figure}
				\centering
				\begin{subfigure}{0.49\textwidth}
					\centering
					\includegraphics[width=0.9\linewidth]{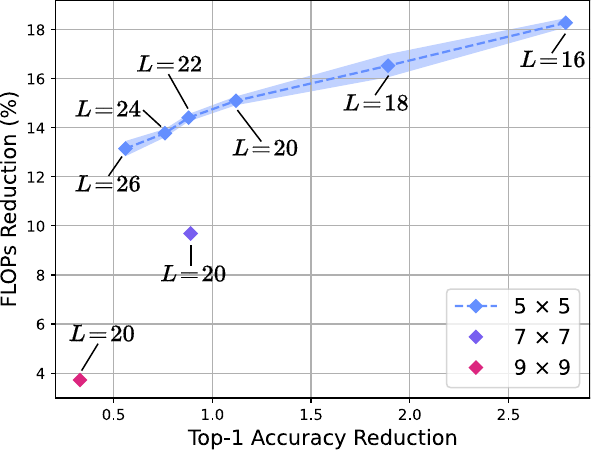}
					\caption{Influence of patch size on FLOPs}
					\label{fig:patch_size_ablation:subfig:tradeoff_flops}
				\end{subfigure}
				\hfill
				\begin{subfigure}{0.49\textwidth}
					\centering
					\includegraphics[width=0.9\linewidth]{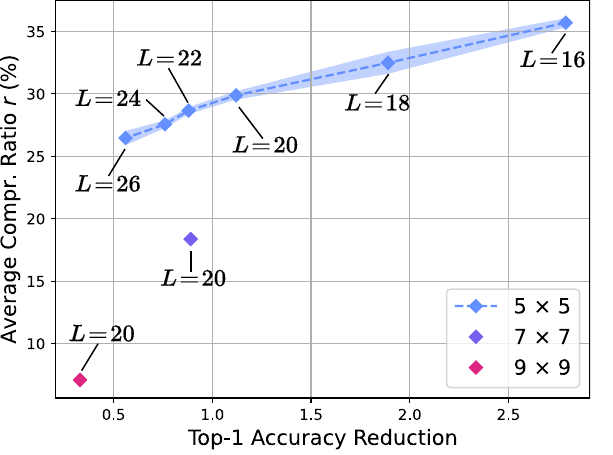}
					\caption{Influence of patch size on compression}
					\label{fig:patch_size_ablation:subfig:tradeoff_compr}
				\end{subfigure}
				\caption{Evaluation of trade-offs between compression and accuracy for different patch sizes. Reproduced from \cite{Meiner2023Instant}}
				\label{fig:patch_size_ablation}
			\end{figure}

			To validate this design, we evaluated alternative patch sizes for a ResNet18 on ImageNet, where the base kernel size is 3$ \times$3. The results, visualized in Figure~\ref{fig:patch_size_ablation}, confirm that our default 5$\times$5 patches provide the best balance of compression and accuracy. Larger patches (7$\times$7, 9$\times$9) disproportionately lose compression efficiency, as depicted in Figure~\ref{fig:patch_size_ablation:subfig:tradeoff_flops}. The best trade-off between accuracy and FLOPs is achieved by using the default patch size of $(K+2) \times (K+2)$ and adjusting the hyperparameter $L$ to the desired point on the Pareto frontier.
			
		\subsubsection{Hyperplane Sparsity}
		\label{sec:experiments:subsec:ablations:subsubsec:hyperplane_sparsity}
			
			The degree of sparsity $s$ in the LSH hyperplanes is a hyperparameter that balances hashing quality and computational cost. We favor sparse, ternary projections over dense Gaussian ones, as they replace expensive floating-point multiplications with simple additions during the hashing step, while offering similar performance \cite{Achlioptas2003Database, Li2006Very}.
			
\begin{figure}
				\centering
				\begin{subfigure}{0.47\textwidth}
					\centering
					\includegraphics[width=0.99\linewidth]{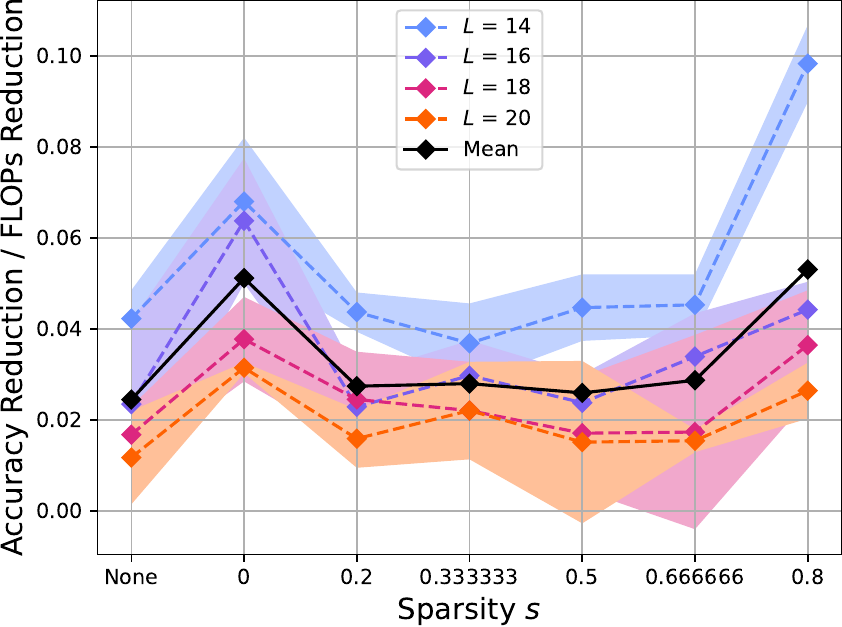}
					\caption{Results for CIFAR-10} 
					\label{fig:sparsity_ablation:subfig:cifar10}
				\end{subfigure}
				\hfill
				\begin{subfigure}{0.47\textwidth}
					\centering
					\includegraphics[width=0.99\linewidth]{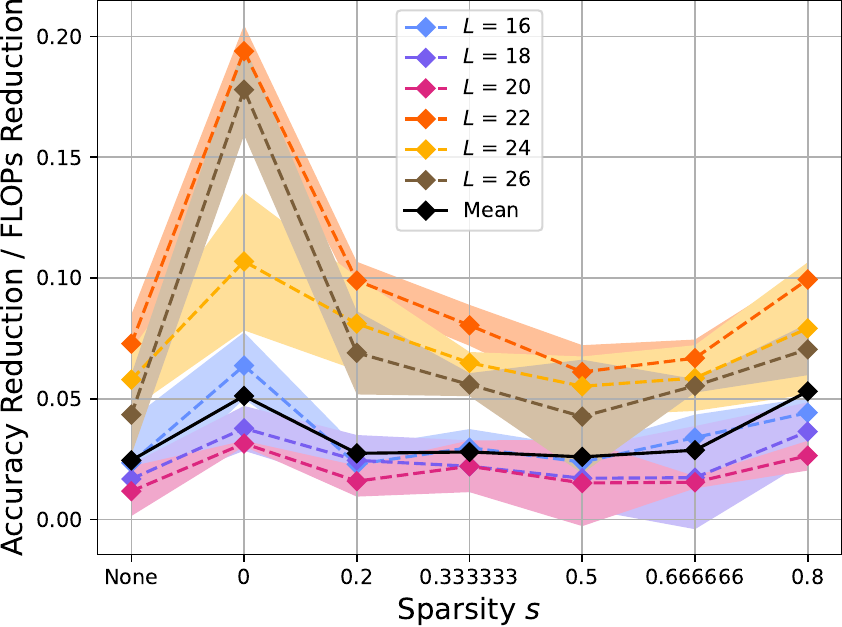}
					\caption{Results for ImageNet} 
					\label{fig:sparsity_ablation:subfig:imagenet}
				\end{subfigure}
				\caption{Influence of the sparsity setting $s$ on different datasets. We plot the ratio of accuracy reduction ($\Delta$) per percentage point of FLOPs reduction for each hyperparameter choice. The setting $s=\text{None}$ denotes the use of dense Gaussian hyperplanes, whereas $s=0$ denotes normal vectors densely filled with entries sampled from $\{-1,1\}$}
				\label{fig:sparsity_ablation}
			\end{figure}

			Our analysis on ResNet18 (Figure~\ref{fig:sparsity_ablation}) shows that the optimal sparsity is primarily dependent on dataset complexity. On CIFAR-10, which is less complex, the model is robust to a wide range of settings for $s$. Since performance is stable even high sparsity, we use $s=2/3$ to maximize computational efficiency. On ImageNet, where latent features are more complex and less redundant, a lower sparsity (i.e., denser hyperplanes) is needed to maintain hashing quality. We found $s=1/2$ to be a robust choice that prevents the performance degradation observed with higher sparsity values.
		
		\subsubsection{Starting Layer Choice}
			The decision of which layer to begin applying \methodname\ is guided by the well-known principle that early network layers learn general, fundamental features, while later layers learn more high-level and often redundant features \cite{Zeiler2014Visualizing}. Compressing early layers therefore tends to negatively impact the model's accuracy.
			
			We evaluate this trade-off on a ResNet18 (see Table~\ref{tab:starting_layer}). As expected, starting the compression at later layers reduces the overall potential for FLOPs reduction. However, it also lessens the impact on accuracy, allowing us to use a less precise hashing scheme (smaller $L$) for the remaining layers, enabling higher compression. Based on this, we adopt a simple, data-dependent policy, as stated in Section~\ref{sec:experiments:subsec:experiment_settings}. For CIFAR-10 models, which exhibit higher redundancy in their features, we begin compression from the first block. For ImageNet models, we begin compression starting from the second layer, preserving the critical early-layer features.
			
			This approach enables high compression while retaining model performance, based on the complexity of the dataset the models were trained on.
			
			\begin{table}
				\caption{Comparison of different starting layer choices for a ResNet18 on CIFAR-10}
				\label{tab:starting_layer}
				\vskip 0.1in
				\centering
				\begin{tabular}{cccccc}
					\toprule
					\multirow[c]{2.5}{*}{\makecell{Starting \\ Layer}} & 
					\multirow[c]{2.5}{*}{$L$} &
					\multicolumn{3}{c}{Top-1 Accuracy (\%)} & 
					\multirow[c]{2.5}{*}{\makecell{FLOPs \\ Reduction (\%)}} \\
					\cmidrule{3-5}
					& & Baseline & Pruned & $\Delta$ & \\
					\midrule
					\multirow[c]{3}{*}{1} & 16 & 93.07 & 91.75 $(\pm 0.24)$ & 1.32 & 38.95 $(\pm 0.37)$ \\
					& 18 & 93.07 & 92.43 $(\pm 0.16)$ & 0.64 & 37.00 $(\pm 0.79)$ \\
					& 20 & 93.07 & 92.52 $(\pm 0.24)$ & 0.55 & 35.73 $(\pm 0.09)$ \\
					\midrule
					\multirow[c]{3}{*}{2} & 14 & 93.07 & 92.32 $(\pm 0.03)$ & 0.75 & 34.00 $(\pm 0.32)$ \\
					& 16 & 93.07 & 92.50 $(\pm 0.06)$ & 0.57 & 32.90 $(\pm 0.17)$ \\
					& 18 & 93.07 & 92.60 $(\pm 0.19)$ & 0.47 & 31.43 $(\pm 0.15)$ \\
					\midrule
					\multirow[c]{3}{*}{3} & 12 & 93.07 & 92.10 $(\pm 0.18)$ & 0.97 & 29.12 $(\pm 0.31)$ \\
					& 14 & 93.07 & 92.44 $(\pm 0.10)$ & 0.63 & 27.99 $(\pm 0.08)$ \\
					& 16 & 93.07 & 92.55 $(\pm 0.09)$ & 0.52 & 27.45 $(\pm 0.31)$ \\
					\midrule
					\multirow[c]{3}{*}{4} & 10 & 93.07 & 93.06 $(\pm 0.01)$ & 0.01 & 17.13 $(\pm 0.04)$ \\
					& 12 & 93.07 & 93.03 $(\pm 0.02)$ & 0.04 & 17.05 $(\pm 0.07)$ \\
					& 14 & 93.07 & 92.96 $(\pm 0.04)$ & 0.11 & 16.90 $(\pm 0.06)$ \\
					\bottomrule 
				\end{tabular}
			\end{table}
		
	\subsection{Performance on Lightweight Architectures}
		\label{sec:experiments:subsec:performance_on_lightweight_architectures}
		
		While our primary focus is on the compression of large, often overparameterized models like ResNets, it is crucial to understand the boundaries of our method's applicability. To this end, we test \methodname\ on MobileNetV2 \cite{Sandler2018MobileNetV2}, an architecture explicitly designed for computational efficiency through the use of depthwise-separable convolutions. Such lightweight models inherently possess less feature redundancy, presenting a challenging test case for our compression scheme.
		
		The results on CIFAR-10, presented in Table~\ref{tab:mobilenetv2}, confirm this hypothesis and highlight the conditions under which \methodname\ operates effectively.
		
		\bmhead{Ineffectiveness of Sparse Hyperplanes} When applying our standard sparse hyperplanes (setting $s=2/3$), the model's retained accuracy collapses. Combined with the high variance, this indicates that sparse hashing does not find truly redundant channels. Compression of these channels then leads to a disruption of the information flow in the MobileNetV2 architecture, lowering its task performance.
		
		\bmhead{Computational Overhead vs. Savings} While switching to dense hyperplanes ($s=\text{None}$) with entries sampled from $\mathcal{N}(0,1)$ mitigates the accuracy loss after compression, it significantly increases the computational overhead of our \methodname\ module. As discussed in Section~\ref{sec:experiments:subsec:results_on_imagenet:subsubsec:impact_of_pointwise_convolutions}, the cost of hashing and merging channels shrinks disproportionately to the cost of the underlying convolution module as the kernel size decreases from 3$\times$3 to 1$\times$1, increasing the relative cost of \methodname. With MobileNetV2's reliance on pointwise convolutions, the baseline cost is already very low. Applying \methodname\ on top and using dense hyperplanes can therefore easily outweigh the savings. 
		\newline 
		
		\noindent We find that a modest benefit is only achieved under a conservative configuration: using dense hyperplanes, only starting the compression scheme in the later layers and using a high number of hash buckets ($L=32$) allows us to save 7.84\% of FLOPs, while losing 1.59 percentage points of accuracy. 
		
		\begin{table}
			\caption{Results for MobileNetV2 on CIFAR-10. "Starting Layer" denotes at which layer we start replacing the convolution modules with our \methodname\ modules. The setting "$s=\text{None}$" denotes the use dense hyperplanes (floating-point values) instead of sparse ones (ternary values). "FLOPs Red." denotes the percentage decrease of FLOPs after compression compared to the base model}
			\label{tab:mobilenetv2}
			\vskip 0.1in
			\centering
			\begin{tabular*}{\linewidth}{@{\extracolsep{\fill}} cccccccc}
				\toprule
				\multirow[c]{2.5}{*}{Model} & 
				\multirow[c]{2.5}{*}{\makecell{Starting \\ Layer}} & 
				\multirow[c]{2.5}{*}{$s$} & 
				\multirow[c]{2.5}{*}{$L$} & 
				\multicolumn{3}{c}{Top-1 Accuracy (\%)} & 
				\multirow[c]{2.5}{*}{FLOPs Red.\ (\%)} \\
				\cmidrule{5-7}
				& & & & Baseline & Pruned & $\Delta$ & \\
				\midrule
				\multirow[c]{13.5}{*}{\rotatebox{90}{MobileNetV2}} & \multirow[c]{6.5}{*}{2} & \multirow[c]{3}{*}{2/3} & 16 & 93.91 & 26.68 $(\pm 5.04)$ & 67.23 & 34.85 $(\pm 0.51)$ \\
				& & & 24 & 93.91 & 70.35 $(\pm 7.28)$ & 23.56 & 26.97 $(\pm 0.25)$ \\
				& & & 32 & 93.91 & 84.33 $(\pm 1.92)$ & 9.58 & 22.67 $(\pm 0.16)$ \\
				\cmidrule{3-8}
				& & \multirow[c]{3}{*}{None} & 16 & 93.91 & 68.45 $(\pm 5.33)$ & 25.46 & 17.38 $(\pm 0.19)$ \\
				& & & 24 & 93.91 & 89.63 $(\pm 0.35)$ & 4.28 & 3.31 $(\pm 0.17)$ \\
				& & & 32 & 93.91 & 92.04 $(\pm 0.16)$ & 1.87 & -7.56 $(\pm 0.25)$ \\
				\cmidrule{2-8}
				& \multirow[c]{6.5}{*}{10} & \multirow[c]{3}{*}{2/3} & 16 & 93.91 & 35.67 $(\pm 8.56)$ & 58.24 & 26.42 $(\pm 0.13)$ \\	
				& & & 24 & 93.91 & 69.91 $(\pm 4.35)$ & 24.00 & 22.08 $(\pm 0.24)$ \\
				& & & 32 & 93.91 & 82.07 $(\pm 2.93)$ & 11.84 & 20.05 $(\pm 0.14)$ \\
				\cmidrule{3-8}
				& & \multirow[c]{3}{*}{None} & 16 & 93.91 & 74.41 $(\pm 2.25)$ & 19.50 & 18.63 $(\pm 0.19)$ \\
				& & & 24 & 93.91 & 90.09 $(\pm 0.57)$ & 3.82 & 12.20 $(\pm 0.16)$ \\
				& & & 32 & 93.91 & 92.32 $(\pm 0.21)$ & 1.59 & 7.84 $(\pm 0.04)$ \\
				\bottomrule 
			\end{tabular*}
		\end{table}

\section{Discussion and Outlook}
\label{sec:discussion}
	\subsection{Method Analysis and Limitations}
		While \methodname\ introduces a new paradigm for training-free dynamic model compression, it is not without limitations. In this section, we aim to provide an overview of the challenges associated with our method and suggest avenues for future work.
		
		\bmhead{Architectural Sensitivity to Pointwise Convolutions} The effectiveness of \methodname\ is reduced on architectures which rely heavily on 1$\times$1 (pointwise) convolutions. As covered in Sections~\ref{sec:experiments:subsec:results_on_imagenet:subsubsec:impact_of_pointwise_convolutions} and \ref{sec:experiments:subsec:performance_on_lightweight_architectures}, the computational overhead of our hashing and merging operations is expensive compared to the very low baseline FLOPs of a 1$\times$1 convolution. This results in diminished net computational savings when applying \methodname. We observed this effect in the performance dip from ResNet34 to ResNet50 on ImageNet (Section~\ref{sec:experiments:subsec:results_on_imagenet:subsubsec:impact_of_pointwise_convolutions}) and in our analysis of MobileNetV2 (Section~\ref{sec:experiments:subsec:performance_on_lightweight_architectures}), where the method struggled to provide a benefit. Consequently, \methodname\ is most effective on standard architectures dominated by larger spatial kernels. Adapting the method to be more efficient for 1$\times$1 convolutions is a key direction for future research, which would unlock its potential for modern lightweight architectures.
		
		\bmhead{Randomness in Hashing} As an LSH-based method, our approach is inherently stochastic. While our experiments consistently show low performance variance across different random seeds, the quality of the channel clustering depends on the randomly initialized hyperplanes. This means that performance can be sensitive to the random seed, particularly at aggressive compression ratios. However, a key advantage of our training-free approach is that exploring multiple seeds is computationally inexpensive. A user can evaluate several seeds in a short time frame to select the best-performing one, mitigating the impact of this randomness in practice.
		
		\bmhead{Common Implementation Challenges} \methodname\ can be categorized as a conditional execution strategy for neural networks, where operations are dynamically selected based on the input. This field of methods faces the challenge that mainstream deep learning frameworks like PyTorch \cite{Paszke2019PyTorch} are heavily optimized for dense, static computations \cite{Belcak2023Exponentially}. Efficiently implementing dynamic or sparse operations often requires custom kernels to achieve theoretical speedups, as standard library functions may not be optimized for such data access patterns. This is not a limitation unique to \methodname, but rather a broader challenge for the research community. The growing body of work on conditional execution \cite{Belcak2023Exponentially, Chen2021MONGOOSE, Kitaev2020Reformer,Belcak2023Fast,Shazeer2017Outrageously,Fedus2022Switch} underscores the need for future software and hardware advancements to support and leverage these methods.

	\subsection{Connection to Transformers: From Channels to Tokens}
		Our work on training-free channel compression in CNNs shares strong conceptual similarities with recent methods for token pruning and merging for Vision Transformers \cite{Dosovitskiy2021Image}, such as EViT \cite{Liang2022Not}, ToMe \cite{Bolya2023Token}, ToFu \cite{Kim2024Token} and other similar works \cite{Yin2022ViT, Rao2021DynamicViT, Chen2023DiffRate, Lee2025Lossless}. At a high level, both strategies utilize the same core principle: identify and compress redundant information in the latent feature representations of a pre-trained model to reduce computational cost, all without requiring retraining to achieve good performance. The primary distinction between both approaches lies in how redundancy is identified and what is ultimately compressed.
		
		In CNNs, similarities between features are not natively computed or utilized by the architecture. Our method therefore introduces an explicit and efficient approximate nearest neighbor search mechanism via locality-sensitive hashing to identify redundant channels. In contrast, ViTs possess a built-in mechanism for discovering feature similarity, namely the self-attention mechanism. Token reduction methods typically leverage this built-in capability, using attention scores \cite{Liang2022Not} or the similarity of key vectors \cite{Bolya2023Token, Kim2024Token} as a signal to identify tokens with redundant information.
		
		The second fundamental difference is the dimension of features which is compressed. \methodname\ operates on the channel dimension of a feature map. It merges similar channels to reduce the feature depth for the convolution operation, but keeps the original output channel dimension. This is necessary to maintain compatibility with the rigid input dimension of subsequent convolutional layers.
		Conversely, token compression operates on the sequence dimension, a flattened representation of the spatial domain. It permanently removes or merges entire tokens from the set, reducing the sequence length for all subsequent transformer blocks. This is possible because the self-attention mechanism and, more generally, the transformer block, is inherently agnostic to the number of input tokens, offering a level of flexibility that CNNs lack.
		
		This analysis raises the question of whether \methodname\ could be extended to ViTs. However, a direct application is not straightforward. Using LSH to find similar tokens would be redundant, as self-attention and its components already provide a more powerful and direct similarity measure.
		
		A more logical extension would be to apply our channel compression scheme to the embedding dimension within a ViT's linear and MLP layers, analogous to our work on CNNs. However, this approach would face the same challenges with relative cost overhead which we identified with 1$\times$1 convolutions (see Sections~\ref{sec:experiments:subsec:results_on_imagenet:subsubsec:impact_of_pointwise_convolutions} and \ref{sec:experiments:subsec:performance_on_lightweight_architectures}). Since each single linear layer operating on tokens is fundamentally equivalent to a 1$\times$1 convolution operating on pixels, the same cost overhead of hashing and merging would apply and likely outweigh the computational savings. This highlights that the working principle of \methodname\ is best suited to computationally expensive operations with inherent redundancy, such as standard spatial convolutions in large CNNs.
		
		This analogy strengthens the emphasis for future work on an efficient extension of \methodname\ to pointwise convolutions. As both linear layers and 1$\times$1 convolutions share the same core principle, such an extension could readily be used in architectures which mostly rely on linear transformations, such as Transformers.

\section{Conclusion}
	In this work, we introduced \methodname, a novel paradigm for dynamic, training-free compression of convolutional neural networks. By leveraging a patch-wise locality-sensitive hashing scheme, our method identifies and merges redundant feature channels at inference time, significantly reducing the computational cost of convolutions without requiring access to any training data. This plug-and-play nature allows for instant compression of publicly available pre-trained models, and even enables changes to the compression ratio in real time.

	Our extensive experiments on CIFAR-10 and ImageNet demonstrate that \methodname\ provides a compelling trade-off between computational savings and model performance across various standard architectures. A key finding is that the effectiveness of our approach scales with model size and width, making it an especially suitable for compressing large, overparameterized models. Furthermore, we conducted a thorough analysis of our method's design principles, including ablation studies on patch size, hyperplane sparsity, and the impact of the starting layer. This analysis also highlighted the method's current limitations, particularly the limited applicability to architectures relying heavily on pointwise convolutions, such as MobileNetV2.
	
	We also contextualized our contributions by drawing parallels to token pruning and merging techniques for Vision Transformers, highlighting both shared principles and unique challenges. Our findings suggest that follow-up work should focus on making \methodname\ more efficient for pointwise convolutions, which would directly unlock the method's potential for a wider range of modern architectures, including Transformers. We hope our work encourages further research into the promising direction of dynamic, training-free model compression that is adaptable at runtime.

\backmatter

\section*{Statements and Declarations}
\bmhead{Funding} This research was funded entirely by Robert Bosch GmbH through an industrial PhD contract for Lukas Meiner. All authors conducted this work as part of their full-time employment at Robert Bosch GmbH. No external funding was received.

\bmhead{Competing Interests} 
All authors are employed by Robert Bosch GmbH. The research presented in this manuscript was conducted for non-commercial, academic purposes, in adherence with the licensing terms of the public datasets used. While the research activity itself is non-commercial, Robert Bosch GmbH may have a future commercial interest in the application of the knowledge gained from this work. The authors declare no other financial or non-financial competing interests.

\bmhead{Author Contribution} All authors contributed to the study's conception and design. The methodology, software implementation, experimental analysis and writing of the original draft were performed by Lukas Meiner. Jens Mehnert and Alexandru Paul Condurache provided supervision and critical review of the work. 

\bmhead{Data Availability}
The datasets and pre-trained models used in the study are publicly available. This work utilizes the CIFAR-10 dataset, available from \url{https://www.cs.toronto.edu/~kriz/cifar.html}, and the ImageNet dataset, available from \url{https://www.image-net.org}. Models pre-trained on CIFAR-10 were sourced from the public GitHub repository \url{https://github.com/huyvnphan/PyTorch_CIFAR10} provided by \citet{Phan2021huyvnphan/PyTorch_CIFAR10}. Models pre-trained on ImageNet are available via the official PyTorch \cite{Paszke2019PyTorch} torchvision library.

\bmhead{Ethics Approval and Consent to Participate} Not applicable.
\bmhead{Consent for Publication} Not applicable.



\end{document}